\documentclass[11pt,a4paper]{article}
\usepackage[hyperref]{emnlp2020}
\usepackage{times}
\usepackage{todonotes}
\usepackage{booktabs}
\usepackage{latexsym}
\usepackage{graphicx}
\usepackage{linguex}
\usepackage{dblfloatfix}
\usepackage{graphicx}

\usepackage{microtype}

\aclfinalcopy 
\def\aclpaperid{3181} 

\newcommand{\change}[1]{{{#1}}}

\title{Investigating representations of verb bias in neural language models}

\author{Robert D. Hawkins$^{*1}$, Takateru Yamakoshi$^{*1,2}$, Thomas L. Griffiths$^1$, Adele E. Goldberg$^1$ \\
$^1$Princeton University, $^2$University of Tokyo \\
\texttt{\{rdhawkins,takateru,tomg,adele\}@princeton.edu} }

\date{}

\begin{document}
\maketitle
\begin{abstract}
Languages typically provide more than one grammatical construction to express certain types of messages. A speaker's choice of construction is known to depend on multiple factors, including the
choice of main verb -- a phenomenon known as \emph{verb bias}.
Here we introduce DAIS, a large benchmark dataset containing 50K human judgments for 5K distinct sentence pairs in the English dative alternation.
This dataset includes 200 unique verbs and systematically varies the definiteness and length of arguments. 
We use this dataset, as well as an existing corpus of naturally occurring data, to evaluate how well recent neural language models capture human preferences.
Results show that larger models perform better than smaller models, and transformer architectures (e.g. GPT-2) tend to out-perform recurrent architectures (e.g. LSTMs) even under comparable parameter and training settings. 
Additional analyses of internal feature representations suggest that transformers may better integrate specific lexical information with grammatical constructions.

\end{abstract}

\section{Introduction}

When we use language, we are often faced with a choice between several possible ways of expressing the same message. For example, in English, to express an event of intended or actual transfer between two animate entities, one option is the \emph{double-object} (DO) construction, in which two noun phrases follow the verb. 
Alternatively, the same content can be expressed using the \emph{prepositional dative} (PO) construction.

\ex. 
\a. Ava gave him something. \hfill \emph{DO}\label{DO}
\b. Ava gave something to him. \hfill \emph{PO}\label{PO}

Speakers' preferences for one or the other construction depend on multiple factors, including the length and definiteness of the arguments  \cite{oehrle1976grammatical, arnold2000heaviness, wasow2002postverbal, bresnan2007syntactic}.
One particularly subtle factor is the lexical \emph{verb bias}.
While some verbs readily occur in either construction, others have strong preferences for one over the other \cite{levin1993english}:

\ex. 
\a. ?Ava said him something. \hfill \emph{DO}\label{DO}
\b. Ava said something to him. \hfill \emph{PO}\label{PO}

\hspace{\parindent} Decades of work in linguistics and psychology has investigated how humans learn these distinctions \cite{gropen1989learnability,perfors2010variability,barak2014learning,goldberg2019explain}.
Yet, as deep neural networks have achieved state-of-the-art performance across many tasks in natural language processing, little is known about the extent to which they have acquired similarly fine-grained preferences.
Although neural language models robustly capture certain types of grammatical constraints, e.g., subject-verb agreement and long distance dependencies \cite{linzen2020syntactic,manning2020emergent}, they continue to struggle with other aspects of syntax, including argument structure \cite[e.g.][]{warstadt2019neural}.
Verb biases provide a particularly interesting testbed. 
Successfully predicting these psycholinguistic phenomena requires the integration of specific lexical information with representations of higher-level grammatical structures, with implications for understanding differential performance between models on other tasks.

\begin{figure*}[!bbp]
    \includegraphics[scale=0.631]{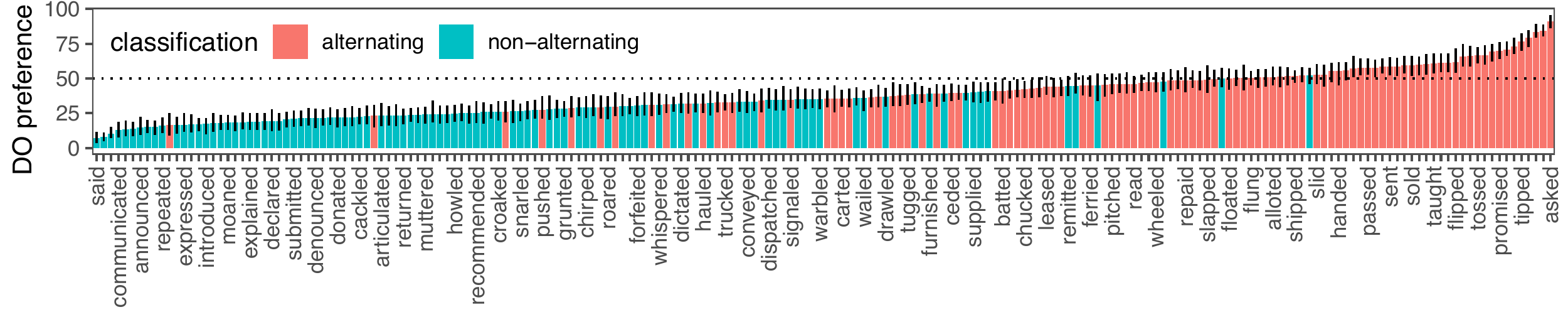}
    \vspace{-1.5em}
    \caption{Human judgments across 200 verbs, pronoun recipients only. Classification from \protect\citet{levin1993english}.}
    \vspace{-1em}
    \label{fig:humanverbs}
\end{figure*}

In the current work, we take an analytic and comparative approach. First, we introduce the \textit{DAIS} (\emph{D}ative \emph{A}lternation and \emph{I}nformation \emph{S}tructure) dataset, containing 50K human preference judgments for 5K sentence pairs, using 200 unique verbs.
These empirical judgments indicate that verb bias preferences are highly gradient in practice \cite{ryskin2017verb,ambridge2018effects}, rather than belonging to binary ``alternating'' and ``non-alternating'' classes, as commonly assumed.
Second, we evaluate the predictions of a variety of neural models, including both recurrent architectures and transformers, and analyze their internal states to understand what drives differences in performance. 
\change{Finally, we evaluate our models on natural production data from the Switchboard corpus, finding that transformers achieve similar classification accuracy as prior work using hand-annotated features \cite[$\sim 93\%$;][]{bresnan2007predicting}.} 

\section{Related Work}

Several recent studies have investigated how neural language models represent the dative alternation.
\citet{kann2018verb} constructed a corpus of verbs in common alternations, including the dative, and showed that a degree of information about acceptability is decodable directly from embeddings of the verb. 
However, acceptability was not based on empirical data and verb bias was treated as a binary variable, preventing an analysis of gradient effects. 
\change{\citet{kelly2020sentence} found that DO constructions are separable from non-DO constructions in high-dimensional sentence embeddings (including BERT), but did not investigate verb bias.}
\citet{futrell-levy-2019-rnns} confirmed that recurrent neural networks (RNNs) show human-like sensitivity to several other important aspects of gradience in dative alternations, including the length and definiteness of arguments.
However, they included only 16 verbs, all considered ``alternating.''
Additionally, in these studies, a limited range of neural models were considered, leaving it unclear exactly how predictions may depend on architectural choices, model size, and training regime.

\section{The DAIS dataset}

The DAIS dataset contains 50,136 human preference judgments for 5,000 sentence pairs, constructed as follows.
First, to obtain a large and heterogeneous set of verbs, we collected the 100 most frequent verbs influentially classified by \citet{levin1993english} as alternating (i.e. acceptably appearing in both PO and DO constructions), as well as the 100 most frequent verbs classified as ``non-alternating'' (appearing only in the PO construction).
This set contains most of the verbs examined in prior corpus-based analyses (see Sec.~5).
For each verb, we generated DO and PO sentences across 5 different conditions, manipulating the length and definiteness of the recipient argument (see ex. 3). 

\ex. 
\a. Ava gave \emph{him} a book. 
\b. Ava gave \emph{the man} a book.
\c. Ava gave \emph{a man} a book. 
\d. Ava gave \emph{the man from work} a book. 
\e. Ava gave \emph{a man from work} a book. 

Finally, to obtain a range of distinct items in each condition, we created 5 plausible theme arguments for each verb, including 2 definite and 3 indefinite, for a total of 5,000 alternation pairs.

We collected judgments from 1011 participants on Amazon Mechanical Turk.
Each participant was shown 50 dative alternation pairs (DO vs. PO) using unique verbs, balanced across the possible recipient and theme conditions. 
On each trial, participants used a continuous slider to indicate the strength of their preference for the DO or the PO, with the midpoint used to indicate they were ``about the same'' (see Appendix A for details)\footnote{ Our procedure and behavioral analysis plan were pre-registered at \textbf{\url{https://osf.io/rtzv4}} and we have released all data and analysis code at \textbf{\url{https://github.com/taka-yamakoshi/neural_constructions}}.}.

\section{Results}
\begin{figure*}[!t]
    \includegraphics[scale=.93]{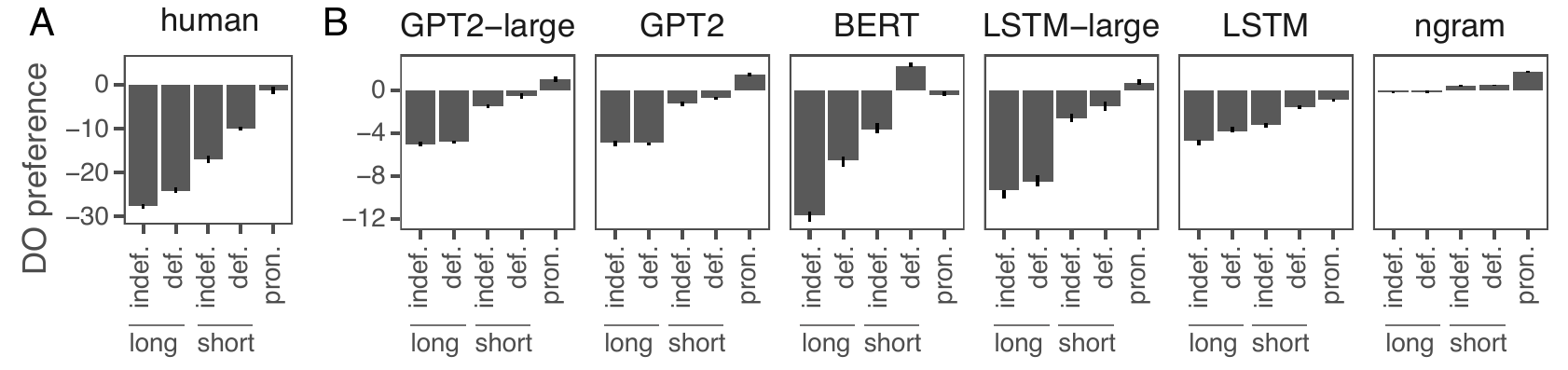}
    \vspace{-1em}
    \caption{Average (A) human and (B) model DO preferences for ``alternating'' verbs, as recipient argument varies in length and definiteness.  For ``non-alternating'' verbs and effects of theme definiteness and, see Fig.~S1 in Appendix.}
    \label{fig:condition_eval}
\end{figure*}

\subsection{Characterizing human judgments}

We begin by characterizing benchmark patterns of human judgments in our dataset. 
First, we examine the degree of gradience in DO preference across verbs.
Traditionally, verbs have been grouped into binary ``classes": alternating verbs which appear freely in both constructions, and non-alternating verbs which are only acceptable in one \cite{levin1993english}.
While verbs in the ``alternating'' class were indeed rated more acceptable on average in the DO than ``non-alternating'' verbs ($b=-15.0, t=-46.5, p<0.001$), there was substantial overlap between the two classes, \change{confirming the need to verify introspective classifications with human judgments.}
Moreover, we found that verbs fell along a continuous spectrum of acceptability in the DO \cite[][see Fig.~\ref{fig:humanverbs}; individual responses are shown in Supplemental Fig.~S2]{lau2017grammaticality,gibson2013need}.
To measure the stability of this ranking across participants, we repeatedly split the dataset in half, measured the mean judgment for each verb across recipient conditions, and took the Spearman correlation between the two halves.
Across 100 splits, we found an average correlation of $r=0.95$, which also serves as a noise ceiling for our model comparison.\footnote{\change{One concern is that gradience is an artifact of using a continuous slider \cite{armstrong1983some,yang2008great}. 
Recent work \cite[e.g.][]{lau2017grammaticality} has addressed this concern by examining the histograms of ratings on different measures, finding higher similarity to gradient control tasks than binary control tasks.
Still, it is important to note that gradient judgments are compatible with categorical grammars due to multiple binary factors or individual differences \cite{schutze2011linguistic}.}}

\begin{table}[b!]
\scalebox{.7}{
\begin{tabular}{c|p{1cm}|p{1cm}|p{1cm}|p{4cm}}
    & \# \,\,\,Layers & Hidden dim. & \# params  & Data (\# tokens) \\ \hline
Ngram&-&-&-&English Wikipedia subset (80M)\\
LSTM&2&650&0.17M&English Wikipedia subset (90M) \\
LSTM-large&2&1024&1.04B&One Billion Word Benchmark(800M)\\
BERT&12&768&110M&BooksCorpus (800M) and English Wikipedia(2.5B)\\
GPT2&12&768&117M& WebText(8B in estimate)\\
GPT2-large&36&1280&774M& WebText(8B in estimate)\\ 
\end{tabular}}
\caption{Details of each model we consider.}
\label{tab:models}
\vspace{-1em}
\end{table}

Second, we examine human sensitivity to the length and definiteness of the arguments (Fig.~\ref{fig:condition_eval}A).
Consistent with previous findings \cite{wasow2002postverbal,futrell-levy-2019-rnns}, participants more strongly preferred the double-object construction when the recipient was shorter ($b= 16.3, t=21.1, p<0.001$) and definite ($b=3.9, t=14.4,p<0.001$), and when the theme was indefinite ($b=2.2, t = 11.0, p <0.001$; see Appendix B for more details).
These effects were roughly additive: although longer recipient arguments rarely occur in the DO construction, we nonetheless found a preference for long definite arguments compared to long indefinites.
Similar effects were found when limiting analysis to only ``non-alternating'' verbs (see Fig.~S1 in Appendix). 
Indeed, ``non-alternating" verbs with short, pronoun recipients were judged to be more acceptable in the double-object construction than ``alternating" verbs with long, indefinite recipients, highlighting the interplay between verb biases and information structure.

\subsection{Comparing model predictions}

\begin{figure*}[t!]
\centering
    \includegraphics[scale=0.78]{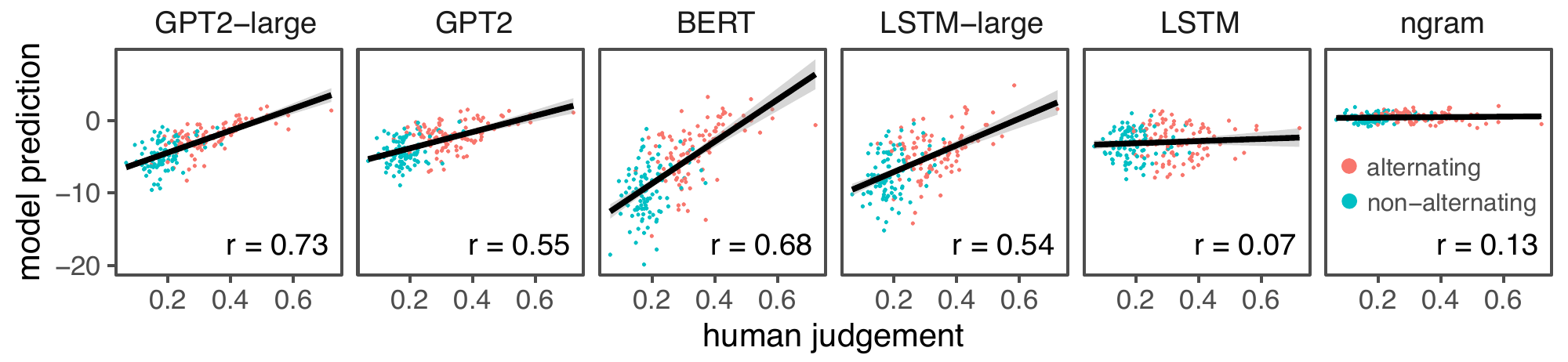}
    \vspace{-1em}
    \caption{Spearman correlation between human judgment and model prediction, across 200 verbs. Judgments were averaged across themes and recipients, hence the lower overall preferences for the double-object.}
        \vspace{-1em}
    \label{fig:humanmodelscatter}
\end{figure*}

Next, we evaluated the performance of several pre-trained neural language models (see Table~\ref{tab:models}) against the fine-grained human judgments in DAIS.
We included two recurrent architectures of different sizes: the 2-layer LSTM model from \citet{gulordava2018colorless}, which has been used for a variety of previous syntactic evaluations, as well as the larger 1B-parameter ``BIG LSTM+CNN" \cite{jozefowicz2016exploring}.
We also included several transformer architectures, including BERT \cite{devlin2018bert}, and two sizes of GPT-2 \cite{radford2019language}.
These choices allow us to explore both effects of architecture as well as size and training regime.
As a baseline, we included a 5-gram model and interpolated using the methods described in \citet{heafield2013scalable}.

For the LSTM and GPT2 architectures, we calculated sentence probabilities by taking the sum of the surprisal of each word, conditioning on all of the preceding words. 
For BERT, which uses bi-directional context, we used the surprisal of each word conditioned on the full context \cite{wang2019bert}.  
We then measured the models' relative preference for the DO construction by taking the likelihood ratio of the two sentences.

We began by examining how each model captures the full spectrum of human verb biases (Fig.~\ref{fig:humanmodelscatter}). 
To do this, we measured the Spearman correlation between human judgments and model predictions across the 200 verbs, averaging over recipients and themes.
We found that the transformer architectures are particularly sensitive to human-like verb biases, with the larger GPT2 model having the highest correlation ($\textit{r} = 0.73$). 
The larger LSTM model had an even greater number of parameters but accounted for significantly less variance, suggesting that simply increasing model size may not be sufficient to learn verb-specific preferences \cite{van2019quantity}.

Next, we examined the extent to which each model qualitatively accounts for human sensitivity to argument length and definiteness (Fig. \ref{fig:condition_eval}B), averaging across verbs. 
For all models except the n-gram model, we found significant effects of recipient length, recipient definiteness, and theme definiteness (see Table S5 in Appendices for details). 
Overall, however, the LSTM models were more sensitive to the effect of definiteness, showing the same additive effects as human speakers. 
Additionally, all models except BERT reflected the fact that ratings on the DO are highest when the recipient is labeled by a (definite) pronoun.

\subsection{Probing internal representations}

Having established key differences in the predictive accuracy of different models, we now investigate the internal representation of this knowledge. 
We hypothesized that sensitivity to verb bias requires the ability to integrate the verb's lexical embedding with the higher-level structure of the sentence. 
Thus, successful models should contain information about acceptability early in the sentence.

To focus on verb bias, we began with the subset of 1000 sentences with pronoun recipients.
For the 4 auto-regressive models (two different sizes of LSTMs and GPT-2), we then extracted the hidden state after each word.
To analyze how acceptability was represented throughout the sentence, we fit regularized linear regressions using the hidden state features as input and human judgments as output (see Appendix C for more details).
We then compared these predictions at three key points in the sentence: after the verb, after the first argument, and after the second argument.

\begin{figure}[!b]
\vspace{-1em}
    \includegraphics[scale=.6]{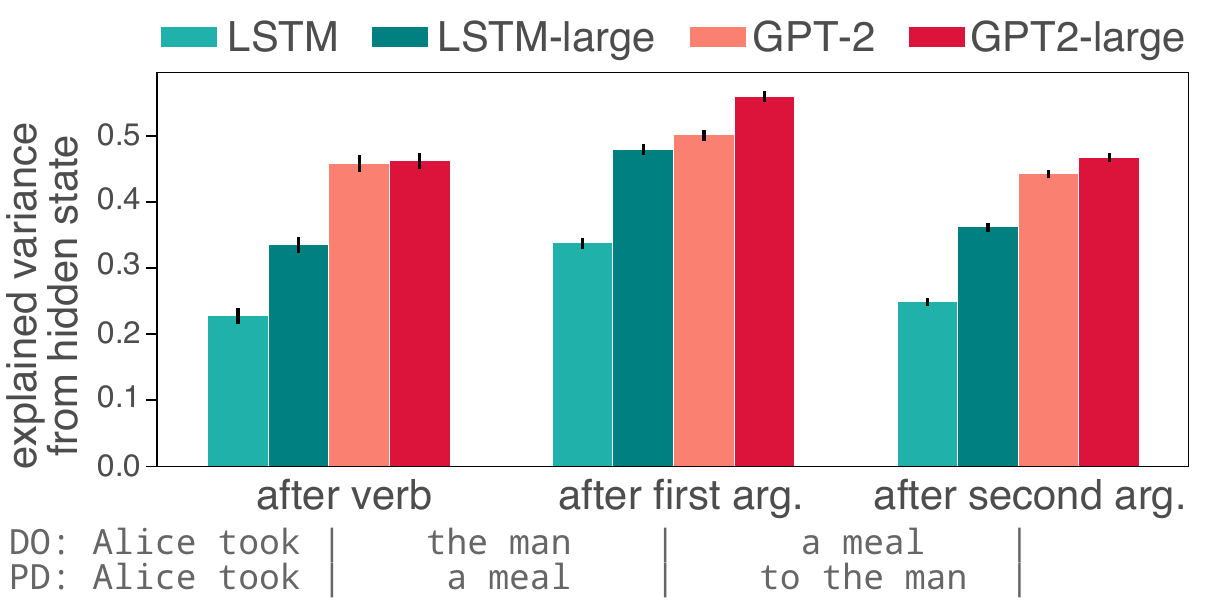}
    \caption{Variance predicted by hidden states throughout the sentence. Error bars show cross-validated SEM.}
    \label{fig:analysis}
\end{figure}

Upon seeing the verb, human preferences for the DO were already decodable from the GPT2 models' features with higher precision than from the LSTMs' (see Fig. \ref{fig:analysis}), reflecting richer lexical representations. 
At the same time, predictive accuracy increased for all models after the first argument, reflecting additional cues from the word sequence.
For example, the model may represent that a pronoun recipient appearing after a PO-biased verb is likely to be less acceptable (e.g. \textit{*Alice said him...}).
Finally, we observed that all of the models lost information about construction preference near the end of the sentence.

\begin{figure}[!t]
    \includegraphics[scale=.5]{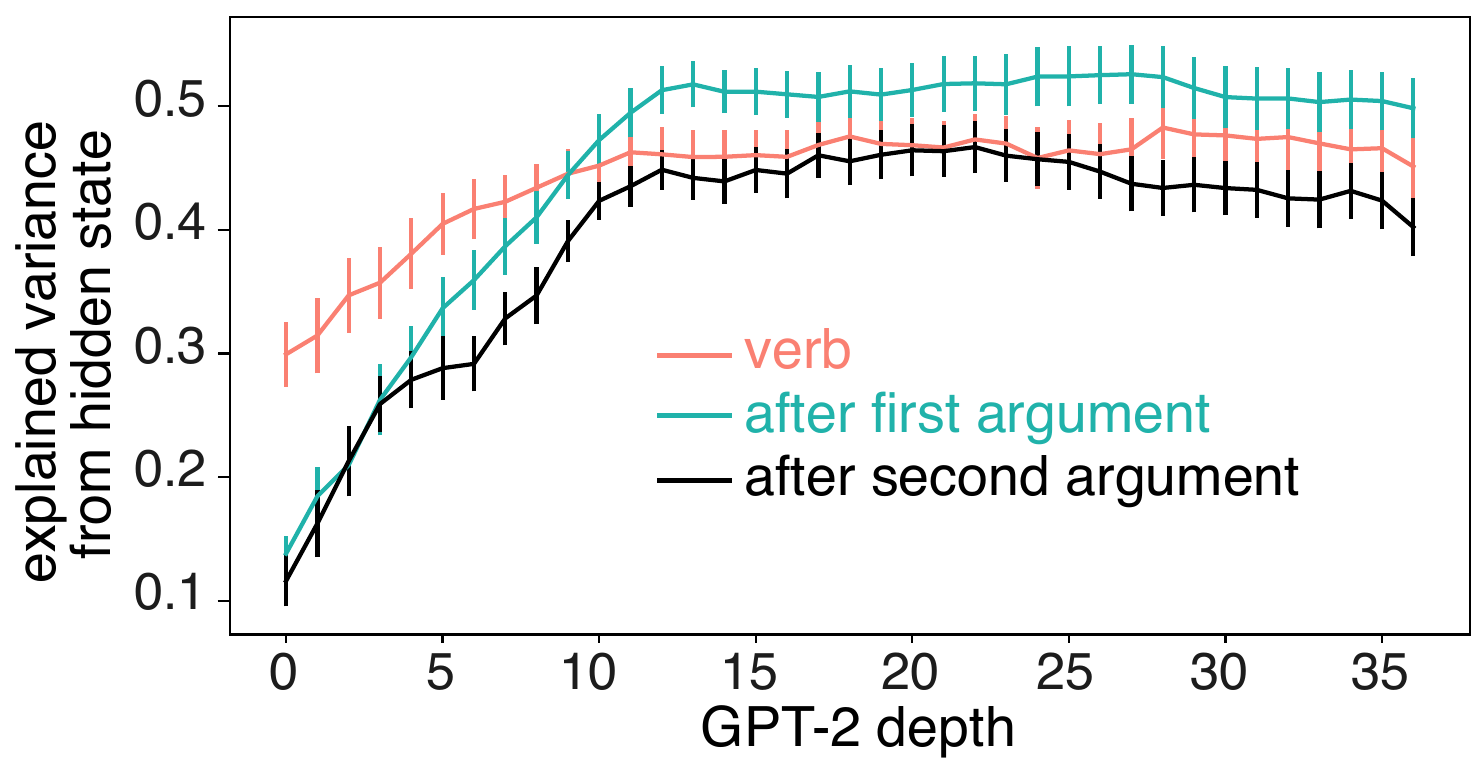}
    \caption{Decodability at each layer of GPT2-large.}
    \vspace{-1em}
    \label{fig:layers}
\end{figure}

How does the representation of verb bias change as a function of depth in the best-fitting GPT2-large architecture? 
Recent analyses have found that the ability to decode syntactic information peaks near the middle layers of transformers \cite{tenney2019bert, hewitt2019structural}.
In the previous analysis, we took the single layer that maximized explained variance; here, we repeat this analysis across all layers (Fig.~\ref{fig:layers}).
Immediately after observing the verb, decodability of DO preferences is already high in the earliest layers, suggesting that this information is directly available from the verb's lexical embedding.
Later in the sentence, however, DO preferences are no longer decodable at lower layers; instead, it has shifted to intermediate layers, suggesting increasing reliance on context and higher-order structure. 

\section{Analysis on natural corpus}

While there are many advantages of human judgment datasets like DAIS --- including the ability to include a wider range of infrequently observed verbs and to control for potential confounds --- there are also distinct advantages of corpus data.
We thus conducted a further evaluation using DO and PO utterances extracted from the Switchboard corpus by \citet{bresnan2007predicting}.
Instead of testing how well each model was able to predict continuous judgments, we now ask how well each model is able to categorically predict whether the DO vs. PO was naturally produced by a corpus speaker.

For each DO or PO utterance in the corpus, we used the extracted verb, theme, and recipient to generate the alternating sentence, pairing the attested example with its hypothetical alternation. 
Subjects were chosen from a list of names, as in the DAIS dataset.
After removing incoherent sentences, we obtained 2,206 pairs of sentences in total.

\begin{table}[b!]
\begin{center}
\normalsize
\begin{tabular}{p{.9cm}|p{.75cm}|p{.8cm}|p{.95cm}|p{.9cm}|p{.9cm}}
GPT2 (large) & GPT2 & BERT  & LSTM (large) & LSTM & Ngram\\ \hline
\textbf{93.51}&93.47&91.29&88.21&81.13&80.05\\
\end{tabular}
\vspace{-.5em}
\caption{Classification accuracy on Switchboard}
\label{tab:SWBD}
\end{center}
\vspace{-1em}
\end{table}

For each model, we calculated the likelihood ratio of the PO vs. DO construction for all corpus examples. 
Following \citet{bresnan2007predicting}, we then constructed a classifier to predict which construction was actually produced by fitting a decision threshold for likelihood ratios. 
The accuracy achieved by each model is shown in Table \ref{tab:SWBD}, reproducing roughly the same ranking that we observed for the DAIS dataset.
Critically, the GPT2 models achieved comparable accuracy to the 92\% previously reported by \citet{bresnan2007predicting} using a logistic regression on 14 hand-annotated binary features (e.g. animacy, accessibility, definiteness). 

An important difference between Switchboard datives and the DAIS dataset is the set of verbs represented. 
Switchboard only contains 37 distinct verbs compared to our 200, and is heavily skewed by frequency (`give' accounts for 42\% of examples). 
Additionally, while we intentionally included 100 verbs traditionally considered ``non-alternating,'' the 37 verbs in Switchboard are skewed toward ``alternating.'' 
Of the 27 of these appearing in \cite{levin1993english}, all were classified as ``alternating,'' and all but one of the 37 appeared in the double-object construction at least once in the dataset.
These features may help account for why Switchboard was unable to distinguish between our GPT2 models: it may be an easier task than DAIS.

\section{Conclusions}

In natural languages, speakers routinely select one alternative over others to express their intended message.  
These choices are sensitive to many interacting factors, including the choice of the main verb and the length and definiteness of arguments. 
Our new dataset, DAIS, not only offers a higher-resolution window into the richness of human preferences, it also provides a newly powerful benchmark for evaluating and understanding the corresponding sensitivity of language models.
We found that transformer architectures corresponded especially well with human verb bias judgments.

Further work is needed to more precisely determine the source of the architectural differences we observed.
One possibility is that the transformer's self-attention mechanism and layer-wise organization improves its ability to represent lexically-specific structures. 
However, it is also possible that differences are attributable to training data.
Another line of future research is to compare the incremental predictions of neural models to finer-grained eye-tracking evidence during sentence processing of double-object sentences \cite[e.g.][]{filik2004processing}.
As neural language models become more complex, subtler phenomena like verb bias may yield new insights into how lexical and grammatical representations are jointly learned and successfully integrated for language understanding.

\section*{Acknowledgements}

This work was supported by a Ma Huateng Data Driven Science SML award to AEG, NSF grant \#1718550 to TLG, and NSF grant \#1911835 to RDH. This work was developed while TY was visiting Princeton, supported by the Ito foundation USA-FUTI scholarship. 
RDH and TY contributed equally and share joint first authorship. 
We are grateful to Tom Wasow, Joan Bresnan, and Tatiana Nikitina for providing corpus materials and thoughtful comments.

\bibliography{emnlp2020}
\bibliographystyle{acl_natbib}

\renewcommand{\thefigure}{S\arabic{figure}}
\renewcommand{\thetable}{S\arabic{table}}
\setcounter{table}{0}
\setcounter{figure}{0}

\section*{Appendix A: Data collection details}

There are many possible ways of empirically eliciting acceptability judgements \cite{marty2019effect,podesva2014research,langsford2018quantifying}. 
We chose to present pairs of sentences together with a continuous slider to maximize our power to detect gradient preferences.
We generated a sentence pair for each verb-theme item by randomly selecting a subject from a list of 8 names (e.g. Juan, Alice), and selecting recipients from a short list corresponding to the given condition (e.g. ``him,'' ``her,'' or ``them'' for the pronoun condition; ``the man,'' ``the woman,'' ``the team'' for the short definite condition, etc.)
See Table \ref{tbl:examples} for examples.
We implemented our study using jsPsych \cite{de2015jspsych} and paid participants a \$1.00 base pay in addition to an additional \$1.00 completion bonus.

To ensure data quality, we excluded participants who failed an initial comprehension quiz or either of two attention checks where one of the sentences in the pair was randomly scrambled:

\ex. 
\a. The man ate a slice of cake.
\b. The man cake of slice ate a.

We also excluded individual trials with response times of $<3$ seconds, and all trials from participants who responded this quickly for more than a quarter of their responses, since it was not possible to read the sentences in that time.
Due to these exclusions, as well as generic participant dropout on Mechanical Turk, not all sentences received the same number of judgements, but we ensured that at least 5 judgements were collected for each sentence pair.

\section*{Appendix B: Regression specifications}

To evaluate the binary effect of alternating vs. non-alternating verbs in Section 4.1, we constructed a mixed-effects model predicting human preferences including a dummy-coded fixed effect for the ``alternating'' vs. ``non-alternating'' classification from \citet{levin1993english}. 
We also included random intercepts and slopes for each human participant.

\begin{figure}[t!]
\centering
    \includegraphics[scale=0.8]{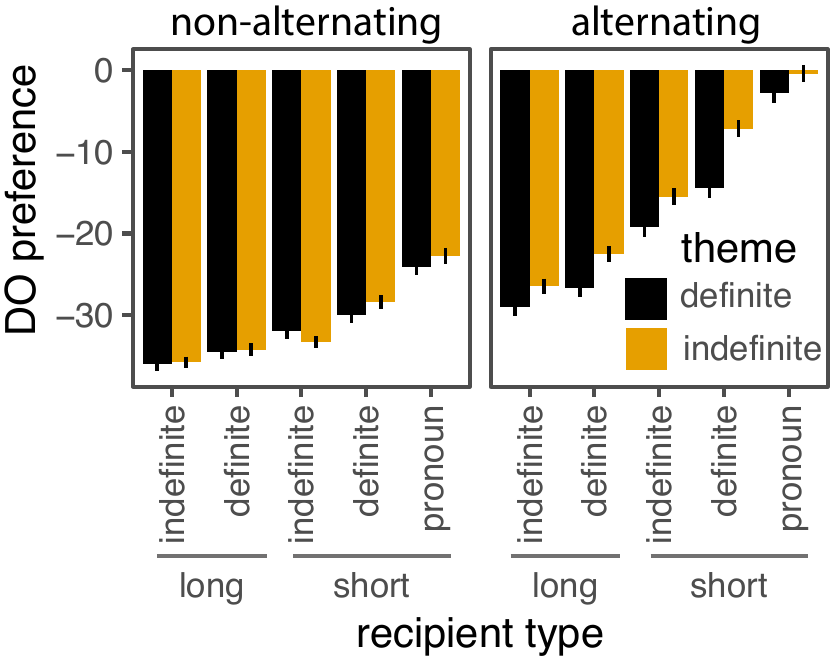}
    \caption{Full pattern of human recipient and theme effects for alternating and non-alternating verbs.}
    \label{fig:appendix}
\end{figure}

To evaluate the effect of information structure in Section 4.1, our mixed-effects model included fixed effects for recipient length, recipient definiteness, and theme definiteness. 
We included random intercepts and effects of recipient length and definiteness for each participant and verb to control for clustered variance at these levels. 
See Fig. \ref{fig:appendix} for the full pattern of results, split by ``alternating'' and ``non-alternating'' verbs.
Complete regression results are shown in Tables S3 and S4.

\section*{Appendix C: Analysis details}

For each of three sentence positions of interest investigated in section 5 (after verb, after first argument, and after second argument), we fit a linear regression predicting human judgements from the hidden states.
Because of the high dimensionality of these states, we used ridge regression to prevent overfitting\footnote{We used the \texttt{sciki-learn} implementation.}.
The ridge regression regularization hyper parameter was optimized for each regression model through a log-scale grid search ($\alpha \in [10^0, 10^7]$) on a held-out validation set.
As our evaluation metric, we computed $R^2$, or variance explained.
Results were averaged across 10 runs of cross-validation, using random 80/20 splits (see Table \ref{tab:reg} for best-performing hyperparameter configurations).

Because the predicted judgements were relative preferences between the two sentences, we concatenated the hidden states of the two sentences together as input.
For the 2-layer LSTMs, we used the final hidden state.
For the deeper GPT-2 architectures, which are known to represent different information at different layers, we did not know \emph{a priori} which layer would be most appropriate.
We thus conducted the regression analysis separately for each layer, and reported the highest performance that was achievable by the model across all layers.
In other words, we computed the cross-validated mean performance for each layer and selected the best.
This approach has also been used in other recent work \cite{schrimpf2020artificial}.

\begin{table*}[]
    \centering
    \begin{tabular}{p{6cm}|p{6cm}}
\textbf{DO sentence} & \textbf{PO sentence} \\
\hline
Michael transported her the food & Michael transported the food to her
\\
\hline
Bob recited the woman something & Bob recited something to the woman
\\
\hline
Juan took a woman a gift & Juan took a gift to a woman \\
\hline
Alice supplied the man who was from work the news & Alice supplied the news to the man who was from work
    \end{tabular}
    \caption{Example sentence pairs}
    \label{tbl:examples}
\end{table*}

\begin{figure*}[t!]
\centering
    \includegraphics[scale=0.43]{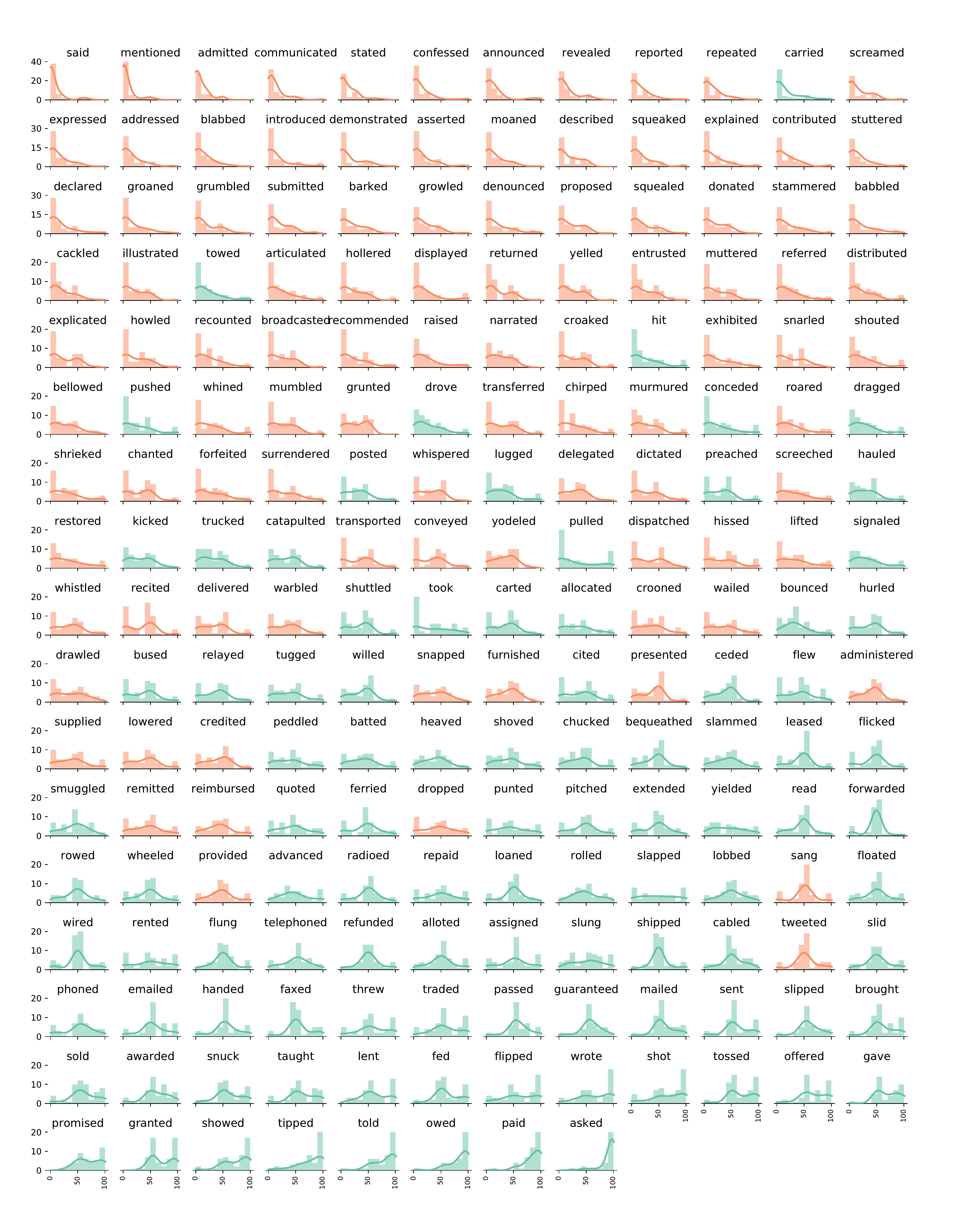}
    \caption{Histograms of individual slider responses for all 200 verbs. Verbs are ranked from lowest mean preference for DO to highest mean preference for DO. Verbs classified as ``non-alternating'' by \citet{levin1993english} colored red, ``alternating'' colored blue.}
    \label{fig:responses}
\end{figure*}

\begin{table*}[!t]
\begin{center}
\begin{tabular}{l|c|c|c|c} 
 &  LSTM  & LSTM-large& GPT2 & GPT2-large \\ \hline
after verb	&1.3($\pm$ 0.2) $\times10^2$ &4.3($\pm$ 0.2)	$\times10^2$	&1.0($\pm$ 0.1)$\times10^4$	&2.4($\pm$ 0.3)$\times10^4$\\
after 1st arg.	&1.1($\pm$ 0.2) $\times10^2$	&2.3($\pm$ 0.2) $\times10^2$	&6.3($\pm$ 0.6)$\times10^3$	& 1.9($\pm$ 0.2)$\times10^4$\\
after 2nd arg.	&1.1($\pm$ 0.2) $\times10^2$	&1.8($\pm$ 0.2) $\times10^2$	&1.9($\pm$ 0.1)$\times10^3$	&3.6($\pm$ 0.4)$\times10^3$\\
\end{tabular}\caption{Regularization hyperparameter configuration for each model and task. SEM across cross-validation runs in parentheses.}
\label{tab:reg}
\end{center}
\end{table*}

\begin{table*}[b!]
\centering
\begin{tabular}{llrrrr} 
\hline
term                                                            & estimate & $t$ statistic & df       & $p$ value                \\ 
\hline
(Intercept)                                                     & 36.44    & 31.02       & 229.52   &  $<1.0 \times 10^{-32}$  \\
recipient length long vs pronoun                                & -16.27   & -21.15      & 257.31   &  $<1.0 \times 10^{-32}$  \\
 recipient length short vs pronoun                               & -8.00    & -18.19      & 281.29   &  $<1.0 \times 10^{-32}$ \\
recipient definite vs. indefinite                               & -3.91    & -14.41      & 194.96   &  $<1.0 \times 10^{-32}$ \\
theme definite vs indefinite                                    & 2.23     & 10.99       & 46616.18 & $<1.0 \times 10^{-32}$ \\
\hline
\end{tabular}
\caption{Fixed effect estimates for human mixed-effects regression, including random effects at the verb-level and participant level. Recipient length, recipient definiteness, and theme definiteness are dummy coded.}
\end{table*}

\begin{table*}
\centering
\begin{tabular}{lllr} 
\hline
random group & term & estimate            \\ 
\hline
participant  & sd(Intercept) & 9.18               \\
participant  & cor(Intercept, recipient length long vs pronoun)                & -0.52 \\
participant  & cor(Intercept, recipient length short vs pronoun)               & -0.45 \\
 participant  & cor(Intercept, recipient definite vs. indefinite)               & -0.76 \\
participant  & sd(recipient length long vs. pronoun)                           & 8.96 \\
 participant  & cor(recipient length long vs pronoun, short vs. pronoun)~       & 0.84     \\
participant  & cor(recipient length long vs pronoun, definite vs indefinite)~  & 0.90 \\
participant  & sd(recipient length short vs. pronoun)                          & 6.81\\
 participant  & cor(recipient length short vs pronoun, definite vs indefinite)~ & 0.91 \\
 participant  & sd(recipient definite vs indefinite)                            & 0.52 \\
 verb         & sd(Intercept)                                                   & 15.70\\
 verb         & cor(Intercept, recipient length long vs pronoun)                & -0.93  \\
 verb         & cor(Intercept, recipient length short vs pronoun)               & -0.76\\
 verb         & cor(Intercept, recipient definite vs.~indefinite)               & -0.80\\
 verb         & sd(recipient length long vs pronoun)                            & 9.22 \\
 verb         & cor(recipient length long vs pronoun, short vs. pronoun)~       & 0.92 \\
 verb         & cor(recipient length long vs pronoun, definite vs indefinite)~  & 0.76 \\
 verb         & sd(recipient length short vs. pronoun)                          & 3.48 \\
 verb         & cor(recipient length short vs pronoun, definite vs indefinite)~ & 0.66 \\
 verb         & sd(recipient definite vs indefinite)                            & 2.19 \\
 Residual     & sd(observation)                                                 & 22.25\\
\hline
\end{tabular}
\caption{Random-effect estimates for mixed-effects regression on human judgments.}
\end{table*}

\begin{table*}
\centering
\begin{tabular}{llrrrll} 
\hline
model      & regression term                             & estimate & $t$ statistic & df      & $p$ value  & sig. level  \\ 
\hline
bert       & (Intercept)                      & -2.83    & -6.66     & 118.43  & 9.22e-10 & ***    \\
bert       & recipient length pronoun vs. long             & -6.08    & -12.75    & 99.34   & 1.23e-22 & ***    \\
bert       & recipient length pronoun vs. short            & 2.59     & 8.00      & 143.07  & 3.91e-13 & ***    \\
bert       & recipient definite vs. indefinite & -5.68    & -18.98    & 99.08   & 9.10e-35 & ***    \\
bert       & theme definite vs. indefinite                  & 4.00     & 19.99     & 2198.00 & 7.95e-82 & ***    \\ 
\hline
gpt2       & (Intercept)                      & 1.01     & 5.59      & 121.11  & 1.43e-07 & ***    \\
gpt2       & recipient length pronoun vs. long              & -6.43    & -29.23    & 100.52  & 6.02e-51 & ***    \\
gpt2       & recipient length pronoun vs. short            & -2.44    & -17.44    & 202.93  & 3.09e-42 & ***    \\
gpt2       & recipient definite vs. indefinite & -0.25    & -2.00     & 99.42   & 4.80e-02 & *      \\
gpt2       & theme\_ypeindef                  & 0.96     & 11.13     & 2198.00 & 5.14e-28 & ***    \\ 
\hline
gpt2-large & (Intercept)                      & 0.20     & 1.09      & 116.31  & 2.80e-01 & n.s.   \\
gpt2-large & recipient length pronoun vs. long             & -5.81    & -27.91    & 99.00   & 1.02e-48 & ***    \\
gpt2-large & recipient length pronoun vs. short            & -1.78    & -12.85    & 99.00   & 7.96e-23 & ***    \\
gpt2-large & recipient definite vs. indefinite & -0.57    & -4.80     & 99.00   & 5.65e-06 & ***    \\
gpt2-large & theme definite vs. indefinite                  & 1.44     & 17.00     & 2099.00 & 8.04e-61 & ***    \\ 
\hline
lstm       & (Intercept)                      & -1.85    & -9.02     & 124.11  & 2.92e-15 & ***    \\
lstm       & recipient length pronoun vs. long              & -2.80    & -8.80     & 100.14  & 4.07e-14 & ***    \\
lstm       & recipient length pronoun vs. short            & -0.87    & -5.26     & 219.65  & 3.44e-07 & ***    \\
lstm       & recipient definite vs. indefinite & -1.33    & -12.04    & 1464.63 & 7.00e-32 & ***    \\
lstm       & theme definite vs. indefinite                  & 1.61     & 16.04     & 2297.00 & 6.16e-55 & ***    \\ 
\hline
lstm-large & (Intercept)                      & -1.19    & -3.05     & 136.46  & 2.74e-03 & **     \\
lstm-large &recipient length pronoun vs. long              & -9.38    & -20.77    & 105.00  & 5.73e-39 & ***    \\
lstm-large & recipient length pronoun vs. short            & -2.30    & -6.98     & 411.84  & 1.16e-11 & ***    \\
lstm-large & recipient definite vs. indefinite & -1.02    & -3.73     & 100.60  & 3.16e-04 & ***    \\
lstm-large & theme definite vs. indefinite                  & 3.21     & 14.67     & 2198.00 & 1.47e-46 & ***    \\ 
\hline
ngram      & (Intercept)                      & 1.27     & 13.27     & 124.45  & 1.39e-25 & ***    \\
ngram      & recipient length pronoun vs. long             & -1.93    & -19.59    & 107.84  & 2.60e-37 & ***    \\
ngram      & recipient length pronoun vs. short            & -1.26    & -12.86    & 107.83  & 1.68e-23 & ***    \\
ngram      & recipient definite vs. indefinite & -0.04    & -0.72     & 98.99   & 4.72e-01 & n.s.   \\
ngram      & theme definite vs. indefinite                  & 0.87     & 16.59     & 2197.99 & 2.59e-58 & ***    \\
\hline
\end{tabular}
\caption{Mixed-effects regression results for each model, including random effects at the verb-level. Recipient length, recipient definiteness, and theme definiteness are dummy coded. *** denotes $p < 0.001$, ** denotes $p < 0.01$, * denotes $p < 0.05$, n.s. denotes `not significant.'}
\end{table*}

\end{document}